\documentclass[pdflatex]{article}
\usepackage{amssymb,amsmath,amsfonts}
\usepackage{scalerel}
\usepackage{graphics,graphicx}
\usepackage[latin1]{inputenc}
\usepackage{yfonts}
\usepackage{proof}
\usepackage{array}
\def \A {{\mathcal{A}}}

\def \C {{\mathcal{C}}}
\def \D {{\mathcal{D}}}
\def \E {{\mathcal{E}}}

\def \K {{\mathcal{K}}}
\def \L {{\mathcal{L}}}
\def \M {{\mathcal{M}}}

\def \o {{\mathcal{O}}}

\def \S {{\mathcal{S}}}
\def \T {{\mathcal{T}}}

\def \W{{\mathcal{W}}}

\def \t {{\mathfrak t}}
\newtheorem{defn}{Definition}

\newtheorem{lemma}{Lemma}
\newtheorem{propn}{Proposition}

\def \lft {\noindent}
\def \ovr {\overline}
\def \otau {\overline{\tau}}

\newcommand{\ignore}[1]{}
\newcommand\Tau{\scalerel*{\tau}{T}\!\!}

\newcommand{\brho}{\mathbf{f}}

\newcommand{\disp}[1]{\vspace*{-1em} \begin{center} {#1} \end{center} \vspace*{-1mm} }

\title{{\Large Distributed Transition Systems with Tags  \\ for Privacy Analysis}}

\author{ \small Siva Anantharaman$^1$ \and
              \small Sabine Frittella$^2$ \and
              \small Benjamin Nguyen$^3$ 
 \ \\ \small
$^1$ LIFO, Universit\'e d'Orl\'eans (France),
            email: {\tt siva@univ-orleans.fr} \\
\small
$^2$ INSA, Centre Val de Loire (France),
email: {\tt sabine.frittella@insa-cvl.fr} \\
 \small
$^3$ INSA, Centre Val de Loire (France),
email: {\tt benjamin.nguyen@insa-cvl.fr} }

 \date{}

\begin{document}

\maketitle

\begin{abstract}
  We present a  logical framework that formally  models how a given {\em private}
  information $P$ stored on a  given database $D$, can get captured progressively,
  by an  agent/adversary  querying  the   database repeatedly.  Named DLTTS 
  ({\em Distributed  Labeled Tagged Transition System}), the framework borrows
  ideas  from  several domains:   Probabilistic Automata  of  Segala, Probabilistic
  Concurrent  Systems,  and Probabilistic  labelled transition systems.
  To every node on a DLTTS is attached a  {\em tag}  that  represents   the `current'
  knowledge of the adversary, acquired from the responses  of  the answering
  mechanism of the DBMS  to his/her queries, at the nodes  traversed  earlier, along
  any given run; this   knowledge is completed at the same node, with further relational
  deductions, possibly  in combination  with  `public'  information from other databases
  given in advance. A `blackbox'  mechanism is  also   part of a DLTTS. It is meant 
  as an oracle, whose role is to  tell  if the  private  information  has been deduced by
  the adversary at the current node, and  if so  terminate the run; an additional special
  feature is that the  blackbox  also  gives   information on how `close', or how `far', the
  knowledge of the adversary is, from  the private information $P$ at the current node.
  A value-wise {\em metric} is  defined for that  purpose,   on the set of  all `type compatible'
  tuples from the   given database, the  data  themselves being  typed with the headers
  of   the  base. Despite the transition systems flavor of our framework, this metric  is
  not `behavioral' in the sense presented in some other works. It is exclusively database
  oriented, and allows to define new   notions of  {\em adjacency} and  of 
  {\em  $\epsilon$-indistinguishabilty}  between   databases, more   generally than those
  usually based on the  Hamming metric with a restricted notion of adjacency. 
  Examples are given all along to illustrate how our framework works. 
  
\ \\[3pt] 
{\em Keywords}: \par Database, Privacy, Transition System, Probability, Distribution.
 \end{abstract}

\section{ { \large Introduction} }

Data anonymization has been investigated for decades, and many privacy
models have been proposed ($k$-anonymity, \emph{differential privacy}, \ldots)
whose goals are to protect sensitive information. 
Our goal in this paper is not to define a new anonymization model, but rather 
to propose a logical framework to formally model how the information stored
in a database can get captured progressively by any agent repeatedly querying
the database. This model can also be used to quantify reidentification attacks
on a database.

The  logical framework we propose below formally  models  how the information
stored in a database can get captured progressively, by an agent/adversary querying
repeatedly the database.  The data can be of any following types: numerical,
non-numerical,or literal. In practice however, some of the literals representing
`sensitive data'  could be in a taxonomical relation; and part of the data could be
presented,  for `anonymization' purposes, as finite intervals or sets, over the basic
types. We shall therefore agree to consider the types of the data in an extended
`overloaded' sense. Cf. Example 1 below.
(Only tree-structured taxonomies will be considered in this work.) 

We assume given a data base $D$, with its attributes set $\A$, usually divided in 
three disjoint groups: the subgroup $\A^{(i)}$ of {\em identifiers}, $\A^{(qi)}$ of
{\em quasi-identifiers}, and  $\A^{(s)}$ of {\em sensitive attributes}. The tuples of
the base $D$ will be generally  denoted as $t$, and their attributes
denoted respectively as $t^{i}, t^{qi},$ and $t^{s}$ in the three subgroups of $\A$.
The attributes $t^{i}$ on any tuple $t$ of $D$  are conveniently viewed  as defining  a
`user' or a `client' of the database $D$.
Quasi-identifiers\footnote{A formal definition of quasi-identifier attributes does
  not seem to  be known.   For our purposes, it suffices to see them as those that
  are not identifiers nor  sensitive.} are  informally defined in general, as a set of public
  attributes, which  in combination with other attributes and/or external information,
  can allow to re-identify all or some of the users to whom the information refers.
The base $D$ itself could be `distributed probabilistically'  over a finite set
(referred to then, as `universe', and its elements named as possible `worlds'). 

By a  {\em privacy  policy} $P = P_A(D)$ on $D$ with respect to a given
agent/adversary $A$ is meant the stipulation that for a certain {\em given set} of
tuples $\{t \in P \subset D\}$, the sensitive attributes $t^{s}$ on any such  $t$ shall 
remain inaccessible (`even after further deduction' -- see below) to  $A$. It is assumed
that $A$ is not the user identified by the attributes $t^{i}$ on these $t$'s. 

The logical framework we propose in this work, to model the evolution of the `knowledge'
that an adversary $A$ can gain by repeatedly querying the given database $D$ -- with a view
to get access to sensitive data meant to  remain hidden for him/her under the given privacy
policy $P$ --, will be called  {\em Distributed Labeled-Tagged Transition  System} (DLTTS);
The underlying logic for DLTTS is first-order, with countably many variables and finitely
many constants (including certain  usual dummy symbols like `$\star, \$, \#$').
In this work, the basic signature $\Sigma$ for the framework is assumed to have no
non-constant function symbols.
By `knowledge' of $A$ we shall mean the data that $A$ retrieves as answers to his/her
successive queries, as well as other data that can be deduced/derived, under relational
operations on these answers; and in addition, some others derivable from these, using
relational combinations with data (possibly involving some of the users of $D$) from
finitely many {\em external  DBs  given in advance}, denoted as $B_1, \dots, B_m$, to
which the adversary $A$ is assumed to have free  access. These relational and querying
  operations are all assumed done with a well-delimited fragment of the language SQL ;
{\em  it is assumed  that this fragment of SQL is part of the signature $\Sigma$ underlying
 the   DLTTSs.} In addition, if $n \ge 1$ is the length of the tuples forming the data in $D$,
finitely many predicate symbols $\K_i, 1 \le i \le n$, each $K_i$ of arity $i$,  will be part
of the signature $\Sigma$; in the work presented here they will be the only predicate
symbols in $\Sigma$. The role of these symbols is  to allow us to see any data tuple of
length $r, 1 \le r \le n$, as a variable-free first-order formula with top symbol $\K_r$,
with all arguments assumed typed implicitly (with the help of the headers of the base
$D$).  In practice however, we shall drop these top symbols $\K_i$, and see
any data tuple that is not part of the given privacy policy $P_A(D)$, directly as a
first-order variable-free formula over $\Sigma$; data tuples $t$ that are elements
of the policy $P_A(D)$ will in practice be just written as $\neg t$.

As we shall see, the DLTTS framework is well suited for capturing the ideas on acquiring
knowledge and on policy violation, in an elegant and abstract setup.  A preliminary
definition of this framework (Section~\ref{DLTTS}) considers only the case where the
data, as well as the answers to the queries, do not involve any notion of `noise'.
(By `noise' we shall mean the perturbation of data by some external random
(probabilistic mechanism.)
But  we shall extend this definition in a later section, as an option to also handle
noisy data.The notion of violation of any given privacy policy on a database can 
then be (optionally)  extended into a notion of violation up to some given
$\epsilon \ge 0$ ($\epsilon$-violation, for short). In the first part of the work, we
will be modeling the lookout for the  sensitive attributes of certain given  users, by
a single adversary. In the second part of the work  (Section~\ref{TwoRuns} onwards),
we propose a method for {\em comparing the evolution of knowledge} of an adversary
at two different nodes on a given run, or on two different possible runs; the  same method
also applies  for comparing the evolution of knowledge  of two different adversaries
$A_1, A_2$, both querying repeatedly (and independently) the given database. 

But before formally defining the DLTTS, a couple of examples might help; they will 
also  throw some light on how to delimit properly the fragment of SQL that we want
included in our logical setup. 

\vspace*{-1em}
\subsection{A couple of Examples}
\vspace*{-1em}

\vspace*{2mm}\lft
 {\bf Example 1}. Table~\ref{1} below is the record kept by the central Hospital of a
 Faculty, with three Departments, in a University, on recent consultations by
 the faculty  staff. In  this record,
 `Name' is an identifier  attribute, `Ailment' is  sensitive, the others are   QI;
 `Ailment' is categorical with 3 branches: Heart-Disease, Cancer,  and
 Viral-Infection; this latter in turn is categorical too, with 2 branches: Flu and CoVid.
 By convention., such taxonomical relations  are assumed known to public, 
 (For simplicity of the example, we assume that all  Faculty staff   are on the
 consultation list of the Hospital.) 

 \begin{table}
   \centering
    \begin{tabular}{|c | c| c| c| c|}
  \hline
  Name  & Age  & Gender & Dept.  & Ailment  \\
  \hline
    Joan    &  24  & F  &  Chemistry  & Heart-Disease  \\
    Michel &  46  & M  &  Chemistry & Cancer  \\
    Aline  &  23  & F  &   Physics     &  Flu    \\
    Harry  &  53  &  M  &  Maths      &  Flu    \\
    John  &  46  &  M  &   Physics    &  CoVid   \\
    \hline
   \end{tabular}
    \caption{\label{1} Hospital's `secret' record }
 \end{table}

 \vspace*{-1mm}
 The Hospital intends to keep  `secret'  information concerning  CoVid infected
faculty members; the tuple  $\neg(John, 46, M, \#, CoVid)$  therefore constitutes
its privacy policy.  The following  Table~\ref{2} is then published for the public,
 where the `Age' attributes have  been anonymized as (integer) intervals,
 the `Ailment' attribute is anonymized by  an upward push in the taxonomy.

 A certain person $A$, who met John at a faculty banquet, suspected John to  have
 been infected with CoVid; (s)he thus decides to consult the published record  of the
 hospital   for  information.  
   \begin{table}[h]
   \centering
    \begin{tabular}{| c| c| c| c|c|}
  \hline
  Age  & Gender &  Dept.  & Ailment  \\
  \hline
    $\ell_1$ & $[20-30[$ & F  &  Chemistry  & Heart-Disease  \\
    $\ell_2$ & $[40-50[$  & M  &  Chemistry & Cancer  \\
    $\ell_3$ & $[20-30[$  & F &   Physics     &  Viral-Infection   \\
    $\ell_4$ & $[50-60[$  & M  &   Maths  &  Viral-Infection \\
    $\ell_5$ & $[40-50[$  &  M  &   Physics   &   Viral-Infection   \\
    \hline
   \end{tabular}
    \caption{\label{2} Hospital's published record }
  \end{table}
   Knowing that the `John'  (s)he met is a `man' and  that the  table~\ref{2} must contain
   John's health bulletin), $A$  has as  choice   lines 2, 4 and 5 ($\ell_2, \ell_4, \ell_5$)
    of  Table~\ref{2}.
   $A$ being in the lookout  for a `CoVid-infected' man, this choice is reduced to the last
   two tuples of the table -- a priori indistinguishable because of the `anomymization'
   (as `Viral-Infection').  Now, $A$ had the impression  that the John (s)he  met `was not too
   old',  so feels that the last tuple is twice more likely; (s)he thus `decides  that John must
   be from the   Physics Dept.', and goes to    consult the  CoVid-cases  statement kept
   publicly visible at that Dept.;  which reads: \par
\hspace*{1cm} Recent CoVid-cases in the Dept:  \;  Female 0  ;  \hspace*{8mm}  Male 1.
\par\lft
And that confirms $A$'s suspicion concerning John.  

In this case, the DLTTS framework would function as follows: At the starting state 
 $s$ a  transition  with three branches would  a priori be possible,  corresponding to
the three  (`M')  lines 2, 4 and 5 of Table~\ref{2},  which represent  the knowledge that
would be acquired  respectively along these branches. Now $A$ is on the lookout for a 
possible CoVid case,  so rules out the `line 2 branch' (i.e., gives this branch probability
 $0$). As for the remaining two branches (corresponding to lines 4 and  5 on
Table~\ref{2}), $A$ chooses to go by the line 5 branch, considering it twice more likely
to be successful, than the other ($A$ had the impression that `John was not too old').
That leads to  the probability distribution $0, 1/3, 2/3$ assigned  respectively on 
the three possible branches for the transition.  If $s_0, s_1, s_2$  are the respective
successor states for the transition considered, the privacy policy of the Hospital
(concerning John's CoVid information) would thus  be  violated at  state $s_2$ (with
probability $2/3$), it wouldn't be at $s_1$ (probability $1/3$);  no information
deduced at state $s_0$.  

  As just seen, modeling an adversary's search for some specific information  on a
 given data base $D$ -- as `runs'  on a suitable DLTTS and  probability distributions  over
 the  successor steps along the runs --, depends in general on the nature(structure)  of the
 information looked for. The probability distributions on the transitions along the runs
 would generally depend  on some  random mechanism, which could also reflect
 the choices  the adversary might make.   \hfill$\Box$

 \vspace*{1mm}
 The role of our next example is to point out that specifying Privacy policy policies 
 will in general have some serious side effects  on the functioning of the primitives and 
 aggregate  procedures of SQL. If the policies are to have some  `content',  operationally
 speaking, the DBMS may have to stipulate that the queries employing these primitives
 either should have `void outputs' in certain contexts, or `get filtered  by the Privacy
 policy'.   
 
\lft {\bf Example 2}. Table~\ref{3} below is an imaginary record $D$ of  a bank $\L$,
containing a  list of  its clients: with client\_ids, their names, and their monthly
balances. (Client\_id  is the  identifier attribute, Monthly--balance is sensitive.)
The privacy policy $P$ of the bank is that client {\em Jean}'s Monthly--balance
should `be invisible'  to  others; formally, the policy $P$ is the negated
formula $\neg (Jean, \ge 420)$. 
  
  On the other hand, the bank is  obliged administratively to render public
  a monthly statement, on its minimum total  Monthly--balance; that is 
  Table~\ref{4}. 
      
  \begin{table}[h]
   \centering
    \begin{tabular}{|c | c| c|}
  \hline
  Client\_id  & Name  & Monthly-balance \\
  \hline
    1  &  Claude  &  320  \\
    2  &   Paul     &  270  \\
    3  &   Jean   &  420   \\
    4  &   Martin  &  150   \\
    5  &   Michel  &  420   \\
    \hline
   \end{tabular}
    \caption{\label{3} $\L$'s (secret) client record }
    
   \vspace*{1em}
  \begin{tabular}{|c | c|}
   \hline
    Number of Clients  &  Minimum Total Monthly--balance \\
    \hline
    5  &  $\ge 1580$  \\
   \hline
   \end{tabular}
  \caption{\label{4} Bank $\L$'s     Monthly public statement} 
  \end{table}

  \vspace*{-1mm}\lft
 An adversary  $A$  wants to know if Jean is a client of the bank, and if 
 so, with a monthly  balance among the   highest.  So $A$ first queries the
 bank to  get  the list of its clients with their  Monthly-balances. The Bank-DBMS's
 answer to $A$'s query will be, say, as in Table~\ref{5} below, where $\star$ stands for
 the anonymization of Jean's sensitive data, as a  `mask' or as an interval,
 say of the form $[330-450[$. 
    
 \begin{table}[h]
   \centering
    \begin{tabular}{| c| c|}
  \hline
  Name  & Monthly-balance \\
  \hline
  Claude  &  320  \\
  Jean   &    $\star$ \\
  Paul     &  270  \\
  Michel  &  420   \\
  Martin  &  150   \\
    \hline
   \end{tabular}
    \caption{\label{5} DBMS's Answer to $A$'s query}
 \end{table}
 
 \vspace*{-1mm}
The external Table~\ref{4} is freely accessible to $A$; so, if the functio\-nalities
{\tt COUNT} and {\tt SUM} are   applied `without any filter',  $A$ can easily deduce that
{\em Jean}'s  Monthly--balance at  $\L$ is $\ge 420$; the bank's Privacy policy is
thus violated.  \hfill $\Box$

{\bf Remark}~1:  In the above example, if the external DB (Table~\ref{4}) was
unavailable to $A$,  the DBMS could have answered his/her query with a Table 5' where
the  entire  tuple on Jean is deleted; in such a case, the privacy policy $P$ on $D$
(concerning  Jean) would a priori remain unviolated; {\em except} if we assume
that the  DBMS accepts queries with aggregate operations on the database $D$ that 
  `do not  explicitly look' for Jean's sensitive attribute: For instance $A$
 could first retrieve the SUM on the entire Monthly--balance column, then ask for
 SUM(Monthly--balance) where `Name $<>$ Jean'.   A relational deduction then
 leads to the violation of the policy $P$.
The above two Examples show  that  the violation of privacy  policies
 needs, in general,  some  additional `outside knowledge'.   \hfill $\Box$

 \vspace*{1mm}
  We may assume  wlog that  the given external bases $B_1, \dots, B_m$ --  to which $A$
 could resort, with relational operations for deducing additional information -- are also
 of the same  signature $\Sigma$ as $D$; so all the knowledge $A$ can deduce/derive
 from his/her repeated queries can be expressed as a first-order variable-free  formula
 over the signature $\Sigma$.  

 \vspace*{-1em}
 \section{{\large Distributed Labeled-Tagged Transition Systems}}~\label{DLTTS}
 \vspace*{-2em}
 
 The DLTTS framework presented in this section synthesizes ideas  coming from 
 various domains, such as the Probabilistic Automata  of Segala (\cite{Segala95b},
 Probabilistic Concurrent  Systems,  Probabilistic labelled transition systems
 (\cite{Fast2018,PTS2019}. Although the  underlying signature for the DLTTS can be
 rich in general, for the purposes of our  current work we shall be working with a
 limited first-order signature (as  mentioned in the Introduction) denoted $\Sigma$,
 with countably many variables, finitely many  constants (including some `standard
 dummies'), no non-constant function symbols,  and  a finite limited set of predicate
 (propositional)  symbols.  Let  $\E$ be the set of all  variable-free formulas  over
 $\Sigma$, and $\mathtt{Ext}$ a given subset of $\E$. We assume given a  decidable
 procedure $\C$ whose role is to `saturate'  any finite set $G$ of  variable-free formulas
 into a finite set $\ovr{G}$, by adding a finite  (possibly empty) set of variable-free 
formulas, using   {\em  relational  operations} on  $G$ and $\mathtt{Ext}$.
This procedure $\C$ will be internal at every node on a DLTTS; in addition, {\em there
will also be a  `blackbox' mechanism $\o$, acting as an  oracle} telling if the given
privacy policy  on a  given database is violated at the current node.
More details will be given in Section~\ref{TwoRuns} on the additional role the oracle
will play in a privacy analysis procedure (for any querying sequence on a given DB),
based on a novel data-based metric, which will be  defined in that section.

\begin{defn}
  A Distributed Labeled-Tagged Transition System (DLTTS), over a given
  signature $\Sigma$, is formed of:
  \begin{itemize}
 \item[-] a  finite (or denumerable) set $S$ of  states, an `initial'  state  $s_0 \in S$,
and a special  state $\otimes \in S$ named `fail':  
 
\item[-] a finite set $Act$ of action symbols (disjoint from $\Sigma$), with  a
  special action $\delta \in Act$ called `violation';
    
\item[-] a (probabilistic) transition relation $\T \subset S\times Act \times Distr(S)$,
  where $Distr(S)$ is the set of all probability distributions over $S$, with finite
  support.

 \item[-] a tag $\tau(s)$ attached to every state  $s \in S\smallsetminus\{\otimes\}$,
  formed of  finitely many first-order variable-free formulas over $\Sigma$;
  the tag $\tau(s_0)$ at the initial state is the singleton set $\{\top\}$.

 \item[-] at every state  $s$ a special action symbol $\iota = \iota_s \in  Act$, 
   said to be  {\em internal}  at $s$, completes/saturates $\tau(s)$ into a set  $\otau(s)$
   with the procedure $\C$, by using relational operations between the formulas in
   $\tau(s)$ and  $\mathtt{Ext}$. 
 \end{itemize}
\end{defn}

\vspace*{-0.5em}
 A (probabilistic) transition $\t \in \T$ will generally be written as a triple 
 $(s, a, \t(s))$; and $\t$ will be said to be `from' (or `at') the state $s$, 
 the states of $\t(s)$ will be the `successors' of $s$ under $\t$.
 The formulas in the tag $\otau(s)$ attached to any  state $s$
 will all be  assigned the same probability as  the state  $s$ in $Distr(S)$. 
 If the set $\otau(s)$ of formulas turns out to be inconsistent,  then the oracle
 mechanism $\o$ will (intervene and) impose $(s, \delta, \otimes)$ as the only
 transition from $s$, standing for `violation'  and `fail',  by definition, 
 
 Nondeterminism of transitions can be  defined without difficulty on DLTTS, as a
 nondeterministic choice between the possible probabilistic transitions at any given
 state.  We shall assume   that nondeterminism is managed by  the  choice of a
 suitable scheduler; and in addition, that at most  one  probabilistic  transition
 is firable  from any state $s \in S\smallsetminus\{\otimes\}$, and none from the
 halting state $\otimes$. 

 \vspace*{1mm}
 \lft {\bf DLTTS and Repeated queries on a database}: 
 The states of the DLTTS will stand for the  various  `moments'  of the querying sequence,
 while the  tags attached to the states  will stand for the knowledge $A$ has acquired
 on the  data of $D$  `thus far'.  This knowledge consists partly in the answers to  the
 queries (s)he  made so far,  then completed with additional knowledge  using the
  internal `sa\-turation' procedure $\C$ of the framework.
 In the context of  DBs, this procedure would consist in relational algebraic
 operations   between the  answers retrieved by $A$ for his/her repeated  queries on
 $D$,  all seen  as  tuples (variable-free formulas), and suitable tuples from the given
 external  databases $B_1, \dots, B_m$.
 If the saturated knowledge of $A$  at a current state $s$ on the DLTTS (i.e., the tag
 $\otau(s)$  attached to the current  state $s$) is not inconsistent, then the  transition
 from  $s$  to  its successor  states  represents the probability distribution of the likely
 answers  $A$  would  expect to get for his/her next query.

 Note that we make no assumption on whether the repeated queries by  $A$ on
 $D$  are treated {\em interactively, or non-interactively}, by the DBMS.  It appears 
 that the logical framework would function exactly alike, in both cases.

 \vspace*{1mm}
 {\bf  Remark}~2: (a) Suppose $\t$ is a  transition from a state $s$, on the DLTTS corresponding
 to a querying sequence by an adversary $A$, and $s'$ is one of the successors of  $s$ under
 $\t$; then, by definition, the `fresh' knowledge  $\tau(s')$ of $A$ at $s'$  resulting from
 this transition, is the addition to $A$'s saturated knowledge  $\otau(s)$ at $s$, the part
 of the response of the DBMS's  answering mechanism for $A$'s  current
 query, represented  by the branch of $\t$ going from $s$ to $s'$.
 
 (b) As already mentioned, we assume that the relational operations needed for  gaining
 further knowledge are  done using a well  delimited finite subset of the functionalities
 of SQL; and   that `no infinite set can get generated from a finite set'  under these 
 functionalities, assumed included in the signature $\Sigma$.   (This corresponds to the 
 {\em bounded inputs outputs} assumption, as in e.g., \cite{BarthePOPL12,BartheLICS20}.)
 \hfill$\Box$
 
  \begin{propn}~\label{exact}
   Suppose given a database $D$, a finite sequence of repeated queries  on $D$ by an 
   adversary $A$, and a first-order relational formula $P = P_A(D)$ over the signature $\Sigma$
   of $D$,  expressing the privacy policy of $D$ with respect to $A$. Let $\W$ be the DLTTS
   modeling the various queries of $A$ on $D$, and the evolution of the knowledge of $A$
   on the data of $D$, resulting from these queries and the internal actions at the states of
   $\W$, as described above.

   (i) The given privacy policy  $P_A(D)$ on $D$ is violated if and only if the failure state
   $\otimes$ on the DLTTS $\W$ is reachable from  the initial state of $\W$.

      (ii) The satisfiabiliity of the set of formulas $\otau(s) \cup \{\neg P\}$ is decidable,  
   at any state $s$ on the DLTTS, under the assumptions of Remark~2(b). 
  \end{propn}
  
  \vspace*{-0.5mm}\lft
  {\em Proof}: Assertion (i) is restatement. Observe now,  that at any state $s$ on
  $\W$,  the tags $\tau(s)$, $\otau(s)$ are both {\em finite sets of first-order variable-free
    formulas} over  $\Sigma$, without non-constant function symbols.  For,   to start
  with, the knowledge of $A$ consists of the  responses received for  his/her queries,
  in the form of a finite set of data tuples from the  given  databases, and some
  subtuples. And by our assumptions of Remark-2~(b), no infinite set can be generated
  by  saturating this  initial knowledge  with procedure $\C$.  
   Assertion (ii) follows then  from the known result that the inconsistency of any given
  finite set of variable-free first-order Datalog formulas  is  decidable, e.g., by the 
analytic tableaux procedure.   (Only the absence of variables is essential.)  \hfill$\Box$

\vspace*{-1em}
 \section{{\large  $\epsilon$-indistinguishability, $\epsilon$-local-differential 
     privacy}}~\label{LDP}
 \vspace*{-2em}
 
 Our objective now is to extend the result of  Proposition~\ref{exact} to the case when 
 the violation to be considered  can be  {\em up to some given $\epsilon \ge0$}, in a
 sense to be made precise.  We stick to the same notation as above. The  set  $\E$ of all
 variable-free formulas over $\Sigma$ is thus a disjoint union of  subsets  of the  form
 $\E = \cup \{\E^{\K}_i \mid 0 < i \le n, \K \in \Sigma\}$, the  index  $i$ in $\E^{\K}_i$
 standing for  the common length of the formulas in the subset, and $\K$  for the
 common  root symbol of its formulas; each  set $\E^{\K}_i$  will be seen as a
 database  of $i$-tuples. 
  
 We shall first look at the situation where the queries intend  to capture  certain
 (sensitive)  values on  a given tuple $t$ in the database $D$.   Two different  tuples
 in  $\E$ might  correspond to two  likely answers  to such a query, but with possibly
 different  probabilities in the distribution assigned for the transitions, by the
 probabilistic mechanism  $\M$ (e.g., as in Example 1).
 
Given two such instances, and  a real $\epsilon \ge 0$, we can also define a  notion
 of their  $\epsilon$-local-indistinguishabilty, wrt the tuple $t$ and  the  mechanism 
 $\M$ answering the queries.
 This can be done in a slightly  extended setup,  where the answering mechanism
 may, {\em as an option}, also add `noise' to  certain numerical data values, for several
 reasons among which the safety of data.   We shall then assume  that the  internal
 procedure $\C$ of the DLTTS  at  each  of its states (meant to  saturate the current 
 knowledge of  the   adversary  querying  the  database) incorporates the following
 three well-known  noise  adding  mechanisms:  the Laplace, Gauss, and  exponential
 mechanisms. With the stipulation that this optional noise additions to  numerical values
 can be done in a {\em bounded} fashion,  so as to be from  a  finite  prescribed domain
 around the values; it will then be assumed that tuples  formed of such noisy data are
 also in $\E$. 

  \vspace*{-1mm}
 \begin{defn}~\label{eps-locindistinguish}
   (i) Suppose that, while answering a given query on the base $D$, at two
   instances  $v, v',$ the probabilistic answering mechanism $\M$ outputs the same
   tuple  $\alpha \in \E$. Given $\epsilon \ge 0$, these two instances are said to be
   $\epsilon$-local-indistinguishable wrt $\alpha$, if and only if: \par   
   \hspace*{2.5cm} $ Prob[\M(v) = \alpha]  \le e^{\epsilon} Prob[\M(v') = \alpha]$ and  \par 
   \hspace*{2.5cm} $ Prob[\M(v') = \alpha]  \le e^{\epsilon} Prob[\M(v) = \alpha]$. 

  (ii) The probabilistic answering mechanism $\M$ is said to  satisfy
   {\em $\epsilon$-local differential privacy} {\em($\epsilon$-LDP)}  for $\epsilon \ge 0$,
   if and only if: For any two instances  $v, v'$ of   $\M$ {\em that lead to  the same
     output},   and any set   $\S \subset Range(\M)$, we have \par 
  \hspace*{2.5cm} $ Prob[\M(v) \in \S ]  \le e^{\epsilon} Prob[\M(v') \in \S]$. 
 \end{defn}

 \vspace*{-1mm}
We shall also be needing  the following notion of  $\epsilon$-indistinguishability (and of 
 $\epsilon$-distinguishability)  of two different outputs of the mechanism $\M$:
 These definitions -- as well that of $\epsilon$-DP given below -- are  essentially
 reformulations of the same (or similar) notions defined in \cite{Dwork2006,Dwork2014}.
 
 \begin{defn}~\label{eps-distinguish}
   Given $\epsilon \ge 0$, two outputs $\alpha, \alpha'$ of  the probabilistic
   mechanism $\M$ answering the queries of an agent $A$, are said to be
   $\epsilon$-indistinguishable,  if and only if:  For every pair   $v, v'$  of inputs
   for $\M$, such that  $Prob[\M(v) = \alpha] = p$ and  $Prob[\M(v') = \alpha'] = p'$,
   we must have:  $p \le  e^{\epsilon} p'$  and  \,$p'  \le  e^{\epsilon} p$.

   \lft
   Otherwise, the outputs  $\alpha, \alpha'$ will be said to be $\epsilon$-distinguishable. 
   \end{defn}

 \vspace*{-1mm}
  {\bf Remark}~3: Given an $\epsilon \ge 0$, one may assume as an option, that 
   at every state on the  DLTTS {\em the retrieval of answers to the current query 
     (from the mechanism $\M$) is done up to $\epsilon$-indistinguishabilty}; this  will
   then be implicitly part of what was called the saturation procedure $\C$ at that state.
   The procedure thus enhanced for saturating the tags at the states, will then be 
   denoted as  $\epsilon\C$, when necessary (it will still be decidable, under the 
   finiteness asumptions of Remark-2~(b)).    Inconsistency of  the set of  formulas,
   in the  `$\epsilon\C$-saturated' tag at any state, will be  checked up to
   $\epsilon$-indistinguishabilty, and referred to as $\epsilon$-inconsistency, or
   $\epsilon$-failure. The notion of   privacy  policy will not need to be modified; that
   of its violation will be referred to as  $\epsilon$-violation,    Under these optional
   extensions of $\epsilon$-failure and $\epsilon$-violation, it must    be clear that  the
   statements  of  Proposition~\ref{exact} continue to be valid.   \hfill$\Box$

   \vspace*{1mm}
  Two small examples of $\epsilon$-local-indistinguishability, before closing this section. 

  (i) The  two sub-tuples ([50--60], M, Maths) and   ([40--50], M,  Physics), from the last
  two tuples on the Hospital's  published record in Example 1  (Table~\ref{2}),   both
  point   to Viral--Infection as output; they can thus be seen as
  $log(2)$-local-indististinguishable, for the  adversary $A$. 

   (ii) The `Randomized Response' mechanism $RR$~(\cite{Warner65}) can be 
  modelled as follows. Input is ($X, F_1, F_2$) where $X$ is a Boolean, and $F_1, F_2$
  are flips of a coin ($H$ or $T$). $RR$ outputs $X$ if $F_1=H$, $True$ if
  $F_1=T$ and $F_2=H$, and $False$ if $F_1=T$ and $F_2=T$. This mechanism is
  $log(3)$-LDP : the instances ($True, H, H$), ($True, H, T$), ($True, T,
  H$) and ($True, T, T$) are $log(3)$-indistinguishable for output $True$.
  ($False, H, H$), ($False, H, T$), ($False, T, H$)and ($False, T, T$) are
  $log(3)$-indistinguishable for output $False$.

 \vspace*{-1em}
 \section{ {\large $\epsilon$-Differential Privacy} }~\label{DP}
  \vspace*{-2em}

  The  notion of  {\em $\epsilon$-indistinguishability  of two given databases} $D, D'$ 
  for an  adversary, is more general than that of   $\epsilon$-local-indistinguishability
  (of pairs of instances  of a probabilistic answering mechanism giving the same
  output, defined in the  previous section).  $\epsilon$-indistinguishability is usually
  defined  only for  pairs of databases $D, D'$ that are  {\em adjacent} in a certain sense
  (cf. below). 

  There is no uniquely defined  notion of adjacence on pairs of databases; in fact, 
  several  are  known, and  in use in the literature. Actually, a notion of adjacence can
  be   defined  in a generic  parametrizable  manner (as in e.g., \cite{dpMetrics2013}),
  as follows.  We assume given a map $\brho$ from the set $\D$ of all databases
  of $m$-tuples  (for some given $m > 0$), into some given metric space $(X, d_X)$.
  The binary  relation  on pairs of databases in $\D$,  defined by  $\brho_{adj}(D, D') =
   d_X(\brho(D), \brho(D'))$   is then said to define a measure of {\em adjacence}  on these 
   databases. The relation $\brho_{adj}$ is said  to define an `adjacency relation'. 
   
  \begin{defn}~\label{eps-indistinguish}
    Let  $\brho_{adj}$ be a given  adjacency  relation on a set $\D$ of databases, and 
    $\M$ a probabilistic mechanism answering  queries on the databases  in $\D$. 
    
    -  Two databases $D, D' \in \D$ are said  to be   $\brho_{adj}$-indistinguishable
    under $\M$,  if and only if, for any   possible output  $\S \subset Range(\M)$,  we  have
     \disp{$ Prob[\M(D) \in \S ]  \le e^{\brho_{adj}(D, D')} Prob[\M(D') \in \S]$.}

     -  The mechanism $\M$ is said to  satisfy     {\em $\brho_{adj}$-differential privacy}
     ($\brho_{adj}$-DP),  if and  only if the above condition is satisfied for  {\em every
     pair of  databases}   $D, D'$ in $\D$, and  any possible output  $\S \subset  Range(\M)$. 
    \end{defn}
      
  \vspace*{-1mm}\lft
    {\em Comments}: (i) Given $\epsilon \ge 0$, the `usual' notions of {\em
      $\epsilon$-indistinguishability and $\epsilon$-DP} correspond to the  choice  of
    adjacency $\brho_{adj} = \epsilon d_h$,  where $d_h$ is the Hamming metric
    on databases -- namely, the number of `records'  where $D$ and $D'$   differ,
    plus the assumption $d_h(D, D') \le 1$ (cf. \cite{dpMetrics2013}). 

    (ii) In Section~\ref{NewDefn}, we propose a more general notion of adjacency,
    based on a different metric  defined `value-wise', to serve other purposes as well. 
    
  (iii) On disjoint databases, one can work  with  different adjacency relations,  using
  different maps to the same (or different) metric  space(s), 

  (iv) The mechanism $RR$  described above is actually $log(3)$-DP,  not only
  $log(3)$-LDP. To check $DP$,   we have to
  check   all possible  pairs of numbers of the form  $(Prob[\M(x) = y], Prob[\M(x') = y])$, 
  $(Prob[\M(x) = y'], Prob[\M(x') = y])$,  $(Prob[\M(x) = y], Prob[\M(x') = y'])$, etc.,
  where   the $x, x'. ...$ are the input instances  for  $RR$, and   $y, y',
  ...$ the outputs.  The mechanism $RR$ has $2^3$ possible input instances for
  ($X, F_1, F_2$) and two  outputs ({\em  True, False}); thus 16
  pairs of numbers, the distinct ones being $(1/4, 1/4), (1/4, 3/4),  (3/4,
  1/4)$, $(3/4, 3/4)$;  if  $(a, b)$ is   any such pair, obviously $a  \le  e^{log(3)} b$.
  Thus $RR$ is indeed $log(3)$-DP.    \hfill$\Box$
 
\vspace*{-1em}
 \section{ {\large Comparing Two Nodes on  one or more Runs} }~\label{TwoRuns}
 \vspace*{-2em}
 
 In the previous two sections, we looked at  the issue  of `quantifying'  the
 indistinguishability of  two data tuples or databases, under repeated queries
 of an adversary $A$. In this section, our concern will be in a sense `orthogonal': 
 the issue will be that of quantifying  how different the  probabilistic  mechanism's
 answers can be, at different moments of $A$'s querying sequence.  Remember that
 the knowledge of $A$, at any node on the DLTTS of the run  corresponding
 to the query sequence, is  represented  as a set of tuples; and  also that the data forming
 any tuple are  assumed implicitly typed, `labeled with' (i.e., under) the headers of the
 database $D$. To be able to compare two tuples of the same length,  we shall
 assume that there is a natural,  injective, {\em type-preserving}  map from
 one of them onto the other; this map will  remain  implicit  in general;  two such  tuples
 will be said to be {\em type-compatible}. If the two tuples are not of  the same length,
 one of them will be projected onto (or restricted to) a suitable  subtuple, so as to 
 be type-compatible and comparable with the other; if this turns out  to be impossible,
 the two tuples will be said to be uncomparable.

 The quantification looked for  will be based on a suitable notion of `distance'
 between  two {\em sets of type-compatible  tuples}. For that, we shall first  define
  `distance' between any two type-compatible  tuples; more precisely, define such 
 a  notion of distance between any two data values under every given header of $D$.
 As a first step, we  shall therefore begin by defining, for every given  header  of  $D$,
 a binary `distance'  function on the set of all values  that get  assigned to the
 {\em  attributes under that header},  along the sequence of $A$'s  queries.
 This distance function  to be defined will be a {\em metric}: non-negative,
 symmetric,  and satisfying the  so-called  Triangle  Inequality (cf. below).
 The `{\em direct-sum}' of these metrics, taken over  all the headers of $D$,
 will then define a metric $d$ on the set of all  type-compatible  tuples  of data
 assigned to the various attributes, under all the headers of $D$, along the
 sequence of  $A$'s queries.
 The `distance'  $d(t, t')$, from any given tuple $t$  in this set to another
 type-compatible  tuple $t'$,  will be  defined as the value of  this direct-sum 
 metric on the pair of tuples $(t, t')$; it will, by definition, be calculated
 `column-wise' on the base $D$, and also on the intermediary databases along
 $A$'s  query sequence; note that it will give us a priori an $m$-tuple of numbers,
 where  $m$ is the  number  of headers (or columns) in the database $D$.
 
 A single number can then be derived  as the  sum of the  entries in the
 $m$-tuple $d(t, t')$. This sum will be denoted as $\ovr{d}(t, t')$, and defined
 as the distance from the tuple $t$ to the tuple $t'$ in the database $D$. 
 Finally, if $S, S'$ are any two given finite sets of type-compatible tuples, of data
 that  get assigned to the various attributes (along the queries), we shall define
 the distance from the set $S$ to the set $S'$ as the number 
 $\rho(S, S') = min \{\, \ovr{d}(t, t') \mid t \in S, \, t' \in S' \, \}$

 \vspace*{1mm}
 Some preliminaries are needed  before we can define the `distance' function between the
 data values  under  every given header of $D$. We begin by dividing the  headers of the
 base $D$ into  four classes classes, for  clarity of presentation:
   
 \vspace*{-1em}
 \begin{itemize}
 \item[.]  `Nominal':  identities, names, attributes  receiving literal data
  {\em not  in any taxonomy} (e.g., gender, city, \dots), finite sets  of  such data;
   \vspace*{-1mm}
 \item[.]  `Numerval' : attributes receiving numerical values, or bounded
   intervals of  (finitely many) numerical values;
  \item[.]   `Numerical': attributes receiving single numerical values (numbers).  
   \vspace*{-1mm}
 \item[.] `Taxoral': attributes receiving literal data in a taxonomy relation. 
\end{itemize}

 \vspace*{-2mm}
  For defining the `distance'  between any two values $v, v'$ assigned to an attribute
 under a given `Nominal'   header of $D$,  for the  sake of uniformity  we agree to 
 consider every value as a  {\em  finite set}  of singleton values.  (In particular, a
 singleton value `$x$' will be seen as the set $\{x\}$.) Given two such values
 $v, v'$, note  first that the so-called {\em Jaccard Index} between them is the number
 $jacc(v, v') = |(v \cap v') / (v \cup v') |$, which is a `measure of their similarity';
 but this index  is not a metric: the {\em triangle inequality} is not satisfied; however,
 the  Jaccard metric $d_{Nom}(v, v') = 1 - jacc(v, v') =  |(v \Delta v') / (v \cup v')|$
 does satisfy that property, and will suit our purposes. Thus defined, $d_{Nom}(v, v')$
 is a `measure of the dissimilarity' between the sets $v$ and $v'$. 
   
  Let  $\Tau_{Nom}$  be the  set of all data assigned to the attributes under the 
  `Nominal' headers of $D$, along the sequence  of $A$'s queries. Then the above
  defined binary  function $d_{Nom}$ extends to a metric on the set   of all
  type-compatible data-tuples from $\Tau_{Nom}$, defined as the `direct-sum' 
  taken over the `Nominal' headers of $D$. 
    
  If $\Tau_{Num}$  is the set of all data assigned to the attributes under  the `Numerval'
  headers along the sequence  of queries by $A$, we also define a `distance'  metric
  $d_{Num}$ on the set of all  type-compatible data-tuples  from $\Tau_{Num}$, in a
  similar manner.   We first define $d_{Num}$ on any couple  of values $u, v$
  assigned to  the attributes under a given `Numerval' header of $D$, then extend it
  to the set of all type-compatible data-tuples from $\Tau_{Num}$ (as the direct-sum
  taken  over the  `Numerval'   headers of $D$).  This will be done exactly as  under
  the `Nominal' headers: suffices to visualize any finite interval value  as a  particular
   way of presenting a set of numerical values (integers, usually). (In particular, a
   single value `$a$'    under a `Numerval'  header will be seen as  the interval value
   $[a]$.) Thus defined the (Jaccard) metric distance $d_{Nom}([a, b], [c, d])$   is a
   measure of `dissimilarity'  between $[a, b]$  and $[c, d]$. . 

   Between numerical data $x, x'$ under the `Numerical' headers, the distance we shall 
   work with is the euclidean  metric $|x - x'|$,  {\em normalized as}: 
   $d_{eucl}(x, x') = |x - x'| / D$, where  $D > 0$ is a fixed finite number,  bigger than
   the maximal euclidean  distance between the numerical data on the databases 
   and on the answers to $A$'s queries. 

   On the data under the `Taxoral'  headers, we choose as distance function  the metric
   $d_{wp}$, defined in  Lemma~\ref{wpmetric} (cf. {\em Appendix}) between the
   nodes of any Taxonomy tree.

   Note that the `datawise  distance functions' defined above  are {\em all with values in
   the real interval} $[0, 1]$.  (This is also one reason for our choice of the distance 
   metric on Taxonomy trees.) This fact is of importance, for comparing  the metric $\rho$
   we defined above with the Hamming metric, cf. Section~ \ref{NewDefn}.

   \vspace*{1mm}\lft
    {\bf An additonal role for  Oracle $\o$}: 
   In Section~\ref{CompScheme} below, we present a  procedure for comparing the
   knowledge of an adversary $A$ at different nodes of the DLTTS that models the
   `distributed  sequence' of $A$'s queries on a given database $D$.    The comparison
   can be  with respect to any given `target' dataset $T$ (e.g., a privacy   policy
   $P$  on $D$).   In operational terms,  so to say, the oracle mechanism $\o$ of 
   the DLTTS  keeps the target dataset `in store'; and as said earlier, a  first role
   for the oracle $\o$ of the DLTTS is to keep a watch on  the  deduction of the target
   dataset by the adversary $A$ at some node. The additional  second role that we
   assign now to the oracle $\o$, is to publish information on the distance  of $A$'s
   saturated knowledge $\otau(s)$,  at any  given node $s$, to the target  dataset $T$.
   This distance is calculated wrt the distance $\rho$, defined above as the minimal
   distance $\ovr{d}(t, t')$ between the tuples $t \in \otau(s), t' \in T$,  where $\ovr{d}$
   is the direct sum of the `column-wise distances'  between the data on the tuples.

   Before presenting the comparison schema, here is an example to  illustrate  how 
   the  notions developed above  operate in practice. 

 \vspace*{1mm}\lft
  {\bf Example 1 bis}. We go back to the Hospital-CoVid example seen earlier,
  more particularly its Table~2, reproduced here: 
 
   \begin{table}[h]
   \centering
    \begin{tabular}{| c| c| c| c|c|}
  \hline
  Age  & Gender &  Dept.  & Ailment  \\
  
  \hline\hline
    $\ell_1$ & $[20-30[$ & F  &  Chemistry  & Heart-Disease  \\
    $\ell_2$ & $[40-50[$  & M  &  Chemistry & Cancer  \\
    $\ell_3$ & $[20-30[$  & F &   Physics     &  Viral-Infection   \\
    $\ell_4$ & $[50-60[$  & M  &   Maths  &  Viral-Infection \\
    $\ell_5$ & $[40-50[$  &  M  &   Physics   &   Viral-Infection   \\
    \hline
   \end{tabular}
   \caption{Hospital's public record recalled}
   \end{table}
  \lft
`Gender' and `Dept.'.  are the `Nominal' headers in this record,  `Age' is `Numerval'
and `Ailment' is `Taxoral'. We are interested in the second, fourth and fifth tuples on
the record, respectively referred to  as $l_2, l_4, l_5$.  The `target set' of
(type-compatible) tuple in this example is taken as the (negation of the) privacy policy
specified, namely the tuple $T = (John, 46, M, \#,  CoVid)$.

We compute now the distance $\ovr{d}$ between the target $T$, and the
three tuples $l_2, l_4, l_5$. This  involves only the subtuple $L = (46, M, \#,  CoVid)$
of $T$: \par
. $\ovr{d}(l_2, L) = d_{Num}(l_2, L) + d_{Nom}(l_2, L) + d_{wp}(L_2, L) $ \par 
   \hspace*{1.5cm}   $ = (1 - 1/10) + 0 + (1 - 2/5) = 9/10 + 3/5 = 15/10 $ \par
. $\ovr{d}(l_4, L) = d_{Num}(l_2, L) + d_{Nom}(l_4, L) + d_{wp}(L_4, L) $ \par 
   \hspace*{1.5cm}  $ = (1 - 0) + 0 + (1 - 4/5) = 1 + 1/5 = 6/5 $ \par
. $\ovr{d}(l_5, L) = d_{Num}(l_5, L) + d_{Nom}(l_5, L) + d_{wp}(L_5, L) $ \par
   \hspace*{1.5cm}  $ = (1-1/10) + 0 +  (1 - 4/5)  =  9/10 + 1/5 = 11/10$ \par
 \lft
 The tuple $l_2$ is the farthest from the target, while  $l_5$ is the closest. This
 `explains'  that the  adversary  can choose the branch on the  transition that leads
 to  a state  where $l_5$ is  added  to his/her knowledge. This is more formally  
 detailed in the procedure presented below. \hfill$\Box$

 \vspace*{-1em}
 \subsection{{\small A (Non-Deterministic) Comparison Procedure}}~\label{CompScheme}
 \vspace*{-1em}
 
\lft
 $\cdot$ Given: DLTTS associated with a querying sequence, by adversary $A$  
 on given database $D$; and {\em a Target set} of tuples $T$.
 
  \lft
  $\cdot$ Given: Two states $s, s'$ on the DLTTS, with respective {\em saturated}
  tags $l, l' $, and probabilties $p, p'$. Target  $T$  assumed not in $l$ or $l'$: neither
  $\rho(l,T)$ nor $\rho(l', T)$ is $0$. Also given: 

   - $config_1$: successor states $s_1, \dots, s_n$ for a transition $\t$ from $s$,
   with probability distribution $p_1, \dots, p_n$; and  respective tags
    $l_1, \dots, l_n$, with the contribution from $\t$ (cf. Remark 2(a)).  

 - $config_2$: successor states  $s'_1, \dots, s'_m$ for a transition $\t'$ from $s'$, 
 with probability distribution $p'_1, \dots, p'_m$; and respective tags
 $l'_1, \dots, l'_m$, with the contribution from $\t'$ (cf. Remark 2(a)). 

 \vspace*{0.6mm}\lft
 $\cdot$ Objective: {\em Choose states to compare under $s, s'$ (with  probability measures
   not lower than $p, p'$) in  $config_1$, or in $config_2$, or from either.}
 
 \vspace*{1mm}\lft
  (i) Compute
  $d_i = \rho(l_i, T), i \in 1 \cdots n$, \,and\,  $d'_j = \rho(l'_j, T), j \in 1 \cdots m$.
 \vspace*{1mm}
\disp{$d_{min}(\t, T) = min\{d_i \mid  i \in 1 \cdots n\}, \, 
               \; \; d'_{min}(\t', T) = min\{d'_j \mid j \in 1 \cdots m\}$} 
   
\lft
(ii)  Check  IF {\em the following conditions are satified} by $config_1$:

    \disp{
    $ d_{min}(\t, T)  \,  \le  \, d'_{min}(\t', T)$ \par 
      $\exists$  an $i, 1 \le i \le n$, such that  $d_i = d_{min}(\t, T)$, $p_i \le p$,  \par 
       and \, $p_i \ge \, p'_j$  for any  $j, \,1 \le j \le m$,  where $d'_j = d'_{min}(\t', T)$ }

  \lft
  (iii)  IF  YES, continue under $s$ with $config_1$, else RETURN. 
   
 \vspace*{-1em}
\section{{\large New Metric for Indistinguishability and DP}}~\label{NewDefn}
\vspace*{-1.5em}

Given a randomized/probabilistic mechanism $\M$ answering the queries on  databases,
and an $\epsilon \ge 0$, recall that the   $\epsilon$-indistinguishability  of any two given
databases under $\M$, and  the notion  of $\epsilon$-DP for  $\M$,  were both defined 
in Definition~\ref{eps-indistinguish} (Section~\ref{DP}), based first on  a
 hypothetical map $\brho$ from the set of all the databases concerned, into some
  given  metric space $(X, d_X)$, and an `adjacency relation' on databases defined as
$ \brho_{adj}(D, D') = d_X(\brho D, \brho D')$,  which was  subsequently instantiated to
  $\brho_{adj} = \epsilon d_h$, where $d_h$ is the Hamming metric between databases.
  It must be observed here, that {\em the Hamming metric is defined only between databases
  with the same number of columns}, and usually only with all  data of the same type.  

  In this subsection, our objective is to propose a more general notion of adjacency, based on
  the distance  metric $\rho$ defined above, between type-compatible tuples on databases
  with   data of multiple types. In other words, our $\D$ here will be  the set of all 
  databases, {\em not necessarily all with the same number of columns, and with data of 
  several possible types} as mentioned in the Introduction. We define then  a binary 
  relation $\brho^{\rho}_{adj}(D, D')$ between $D, D'$ in the set  $\D$ by 
  setting $ \brho^{\rho}_{adj}(D, D') = \rho(D, D')$, visualizing  $D, D'$  as  sets of
  type-compatible data tuples. 
 
  Given $\epsilon$, we can then define the notion of $\epsilon_{\rho}$-indistinguishabilty  of 
  two  databases $D, D'$ under a (probabilistic) answering mechanism $\M$, as well as  the
  notion of $\epsilon_{\rho}$-DP for $\M$, exactly as in Definition~\ref{eps-indistinguish},  by 
  replacing   $\brho_{adj}$ first with the relation $ \brho^{\rho}_{adj}$,  and subsequently  with
  $\epsilon\rho$. The notions thus defined are {\em more general}  than those presented
  earlier in  Section~\ref{DP} with the choice  $\brho_{adj} = \epsilon  d_h$.
 An example will illustrate this point.  
  
  \vspace*{1mm}\lft
 {\bf Example 4}. We go back  to  the `Hospital's public record' of our 
 previous example, with the same notation. For this example,  we shall
 assume that the  mechanism $\M$ answering a  query for `ailment  information
 involving men' on  that record, returns the tuples  $l_2, l_4, l_5$ with the
 probability distribution $0,  2/5,  3/5$, respectively.
Let us  look for the minimum value of $\epsilon \ge 0$, for which these three tuples
 will be $\epsilon_{\rho}$-indistinguishable under the mechanism $\M$.

 The output $l_2$, with probability $0$, will be $\epsilon_{\rho}$-distinguishable
 for any $\epsilon\ge 0$. Only the two other outputs  $l_4, l_5$ need to be considered.
  We first compute the  $\rho$-distances between these two tuples:
 $ \ovr{d}(l_4, l_5) = (1 - \frac{1}{20}) + 0 + 1 + 0 = 39/20$. The condition for $l_4$ and
  $l_5$ to be $\epsilon_{\rho}$-indistinguishable under $\M$ is thus:
  
 \disp{ $ (2/5 ) \le e^{(39/20) \epsilon}*(3/5)  \;\;  and  \;\;  (3/5) \le  e^{(39/20)
     \epsilon}*(2/5)$,}
 i.e.,   $\epsilon \ge (20/39)*ln(3/2)$. 
 In other words, for any $\epsilon \ge (20/39)*ln(3/2)$, the two tuples $l_4$
 and $l_5$ will be $\epsilon_{\rho}$-indistinguishable; and for  values of  $\epsilon$ with
 $0 \le \epsilon <   (20/39)*ln(3/2)$, these tuples will be  $\epsilon_{\rho}$-distinguishable.
 
 For  the $\epsilon$-indis\-tinguishabilty of these tuples wrt the  Hamming metric $d_h$,
 we proceed similarly:  the distance $d_h(l_4, l_5)$ is by definition the number
 of `records' where these tuples differ, so  $d_h(l_4, l_5) = 2$. So the condition on
 $\epsilon \ge 0$  for their $\epsilon$-indistinguishabilty wrt $d_h$ is:
   $(3/5)  \le  e^{2 \epsilon}*(2/5)$,\, i.e., \, $\epsilon \ge (1/2)*ln(3/2)$ .

 In other words, if these two tuples are $\epsilon_{\rho}$-indistinguishables wrt $\rho$
 under $\M$ for some $\epsilon$, then they will  be $\epsilon$-indistinguishable wrt
 $d_h$ for  the same  $\epsilon$. But the converse is not true,  since  
 $(1/2)*ln(3/2) <  (20/39)*ln(3/2)$. Said otherwise: {\em  $\M$   $\epsilon$-distinguishes
  more finely with  $\rho$, than with $d_h$}.  \hfill$\Box$

 \vspace*{1mm}
  {\bf Remark}~4:  The statement!``$\M$  $\epsilon$-distinguishes  more finely with 
  $\rho$, than with $d_h$'',  is {\em always true} (not just in Example~4). For the following
  reasons: The records  that differ `at some  given position' on  two bases  $D, D'$  are
  always at distance $1$ for  the Hamming  metric $d_h$, by definition, whatever be
  the  type of data stored at that position.
  Now, if the  data stored  at that position `happened to be'  numerical, the usual 
  euclidean distance  between the  two data could have been (much) bigger  than
  their Hamming distance $1$; precisely to avoid such  a situation, our definition  of
  the metric $d_{eucl}$ on numerical data `normalized'  the euclidean  distance,
  to ensure that their $d_{eucl}$-distance will not exceed their Hamming distance.
  Thus, all the `record-wise' metrics  we have defined above  have their values in
  $[0,1]$,  as we mentioned earlier; so, whatever the type of data at corresponding
  positions on any two bases $D, D'$, the $\rho$-distance between the records will
  never exceed their Hamming distance. That suffices to prove our  statement above. 
 The Proposition below formulates all this, more precisely:

  \begin{propn}
    Let  $\D_m$ be the set of all databases  with the same number $m$ of columns,
    over a finite set of given data, and $\M$ a probabilistic mechanism answering
    queries on the  bases  in $\D$.  Let $\rho$ be the metric (defined above) and  $d_h$
    the Hamming metric, between the databases in $\D$, and suppose  given
    an $\epsilon \ge 0$. 
    
    - If two databases $D, D' \in \D_m$ are $\epsilon_{\rho}$-indistinguishable under $\M$
    wrt $\rho$, then they are also  $\epsilon$-indistinguishable under $\M$ wrt $d_h$.
   
    - If the mechanism $\M$  is  $\epsilon_{\rho}$-DP on the bases in $\D_m$ (wrt $\rho$),
    then it is also $\epsilon$-DP (wrt $d_h$) on these bases.   
  \end{propn}

  The idea of `normalizing'  the Hamming metric between numerical databases (with the
  same number of columns) was already suggested in   \cite{dpMetrics2013} for the same
  reasons. When only numerical databases are  considered, the metric   $\rho$ that 
  we have defined above  is  the same as the `normalized Hamming   metric' of
  \cite{dpMetrics2013}.  Our metric  $\rho$ must actually be seen as a generalization
  of that  notion, to directly handle  bases  with more general  types of  data: 
  anonymized,  taxonomies, \dots  

 \vspace*{-1em}
 \section{ {\large Related Work and Conclusion} }
  \vspace*{-1mm}
  
A starting point for the work presented is the observation that  databases
  could be distributed over several `worlds' in general, so querying  such bases
  leads to answers which would also be distributed; to such distributed  answers
  one could conceivably assign probability distributions  of relevance  to
  the query. The probabilistic automata of Segala (\cite{Segala95a,Segala95b}) are 
  among the  first logical  structures proposed to model  such a vision, in
  particular  with outputs.  Distributed  Transition Systems (DTS) appeared 
  a little later, with as objective the behavioral  analysis of the  distributed
  transitions, based on  traces or  on  simulation/bisimulation, using quasi- or pseudo-
  or hemi- metrics   as in~\cite{Fast2018,PTS2019,LBr-SystMetrics09}.
  Our  lookout in this work was for a  syntax-based {\em metric in the mathematical
  sense,} that can directly handle  data  of  `mixed' types -- which 
  can be numbers or literals, but can  also be `anonymized'  as intervals or sets;
  they can also be taxonomically  related to each other in a  tree structure.
  (The metric $d_{wp}$ we have defined  in the   {\em Appendix} on the nodes
  of a  taxonomy  tree is novel.)  Data-wise metrics as defined  in  our work   can 
  express more precisely, in a mathematical sense, the `estimation errors' of an
  adversary wrt the given privacy policies on the database, at any point of his/her
  querying process. (In ~\cite{Monedero2013}, such estimations are expressed
  in terms of suitably defined `probability measures'.) 
  Implementation and experimentation are part of  our future work, where we also
  hope to   define a `divergence measure' between two given nodes   on a DLTTS
  modeling a querying process, in terms of the know\-ledge distributions at the
  two nodes -- independently of any notion of a given target data set.

  \section*{ {\large Appendix} }

  Taxonomies are frequent in machine learning.  Data mining  and clustering  techniques
  employ reasonings based on measures of symmetry, or on metrics, depending  on the
  objective. The Wu-Palmer symmetry  measure on tree-structured   taxonomies is
  one among those in use; it is defined as follows (\cite{WuPalm1994}):
  Let $\T$ be a  given  taxonomy tree.   For any node $x$ on $\T$, define its depth
  $c_x$ as the number of nodes  from the root  to $x$ (both included), along the path
  from the root to $x$. For any pair $x, y$ of nodes  on $\T$, let $c_{x y}$ be the depth
  of the common ancestor of $x, y$ that is  {\em farthest } from the root.
  The Wu-Palmer symmetry  measure between the nodes  $x, y$  on $\T$ is then
  defined as WP$(x, y) = \frac{2 \, c_{x y}}{c_x + c_y}$.
  This measure, although considered satisfactory for many purposes, is known to have
  some disadvantages such as not being conform to semantics in several situations.

  What we are interested in, for the purposes of our current paper, is a   {\em metric}
  between the nodes of a taxonomy tree, which in addition will suit our semantic
  considerations.  This is the objective of our Lemma below. (A result that seems to be
  unknown, to our knowledge.) 

  \vspace*{-1mm}
  \begin{lemma}~\label{wpmetric}
    On any taxonomy tree $\T$, the  binary function between its
    nodes defined by \,  $d_{wp}(x, y) = 1 - \frac{2 \, c_{x y}}{c_x + c_y}$ {\em (notation
    as above)} is a metric. 
  \end{lemma}
  
  \vspace*{-1mm}   \lft
  {\em Proof}: We drop the suffix $wp$ for this proof, and just write $d$.
  Clearly $d(x, y) = d(y, x)$; and $d(x, y) = 0$ if and only if $x = y$.  We only have to
  prove the  Triangle Inequality; i.e. show that   $d(x,z) \le d(x, y) + d(y,z)$ holds for
  any three nodes $x,y,z$ on $\T$.  A `configuration'  can be  typically represented
  in its `most  general form'  by the  diagram below.
  The boldface characters   $X, Y, Z, a, h$ in the diagram stand  for the  {\em number
    of arcs}  on the corresponding paths. Thus, for the depths  of the  nodes $x, y, z$,
  and  of their farthest common  ancestors on   $\T$,  we get:  
  \disp{$c_x = X + h + 1 ,  \; \; c_y = Y + h + a + 1,  \;\;  c_z = Z + h + a + 1$, 
    $c_{xy} = h + 1, \;\;   c_{yz} = h + a + 1,  \;\;  c_{xz} = h + 1$} 
  \vspace*{-1mm}
The `$+1$' in these equalities is because the $X, Y, Z, a, h$ stand for the {\em number
of arcs} on the paths, whereas the depths are defined as the number of nodes.
Also note  that the $X, Y, Z, a, h$ must all be  integers $\ge 0$. 

For the Triangle Inequality on  the three nodes $x, y, z$ on $\T$,
 it suffices to prove the following two relations: 
 \disp{$d(x, z) \le d(x, y) + d(y, z)$ \, and \,  $d(y, z) \le d(y, x) + d(x, z)$.}
 \vspace*{-1mm}\lft
 by showing that the following two algebraic inequalities hold: \vspace*{0.5mm}
\disp{\!(1) $1 - \frac{ 2*(h+1) }{ (X+Y+2*h+a+2) } + 1 - \frac{ 2*(h+a+1) }{ (Y+Z+2*h+2*a+2) }$
  $\ge$     $1 - \frac{  2*(h+1) }{ (X+Z+2*h+a+2) }   $ \par
   \, (2) $1 - \frac{ 2*(h+1) }{ (X+Y+2*h+a+2) } + 1 - \frac{ 2*(h+1) }{ (X+Z+2*h+2*a+2) }$
   $\ge$     $1 - \frac{  2*(h+a+1) }{ (Y+Z+2*h+2*a+2) }   $} \vspace*{-0.5mm}
The third  relation $d(x, y) \le  d(x, z) + d(z, y)$ is proved by just
exchanging the roles of $Y$ and $Z$ in the proof  of inequality (1).

 \lft Inequality (1): We eliminate  the denominators (all  strictly positive), and  write
 it out as an inequality between two polynomials  $eq1, eq2$ on $X,Y,Z$, $h,a$, which
 must be satisfied for all their non-negative integer values:

 \vspace*{1mm}
\lft $ eq1: (X+Y+2*h+a+2)*(Y+Z+2*h+2*a+2)*(X+Z+2*h+a+2)$ 
\lft $ eq2:  (h+1)*(Y+Z+2*h+2*a+2)*(X+Z+2*h+a+2) $ \\ \hspace*{1.5cm}
                    $ +(h+a+1)*(X+Y+2*h+a+2)*(X+Z+2*h+a+2)$ \\ \hspace*{1.5cm}
                    $ - (h+1)*(X+Y+2*h+a+2)*(Y+Z+2*h+2*a+2)$  \\ 
$eq: eq1 - 2*eq2$.  \; We need to check:\, $eq \ge 0$ ? 

\vspace*{1mm}\lft
The equation $eq$ once expanded  (e.g., under {\em Maxima})  appears as:

\vspace*{1mm}
\disp{
$eq: YZ^2+XZ^2+aZ^2+Y^2Z+2XYZ+4hYZ+2aYZ+4YZ+X^2Z+4hXZ+2aXZ+4XZ+a^2Z+XY^2
               +4hY^2+aY^2+4Y^2+X^2Y+4hXY+2aXY+4XY+8h^2Y+8ahY+16hY+a^2Y+8aY+8Y$}
\lft
The coefficients are all positive, and inequality (1)  is proved. 

\vspace*{-0.7em}
  \includegraphics[scale=0.6]{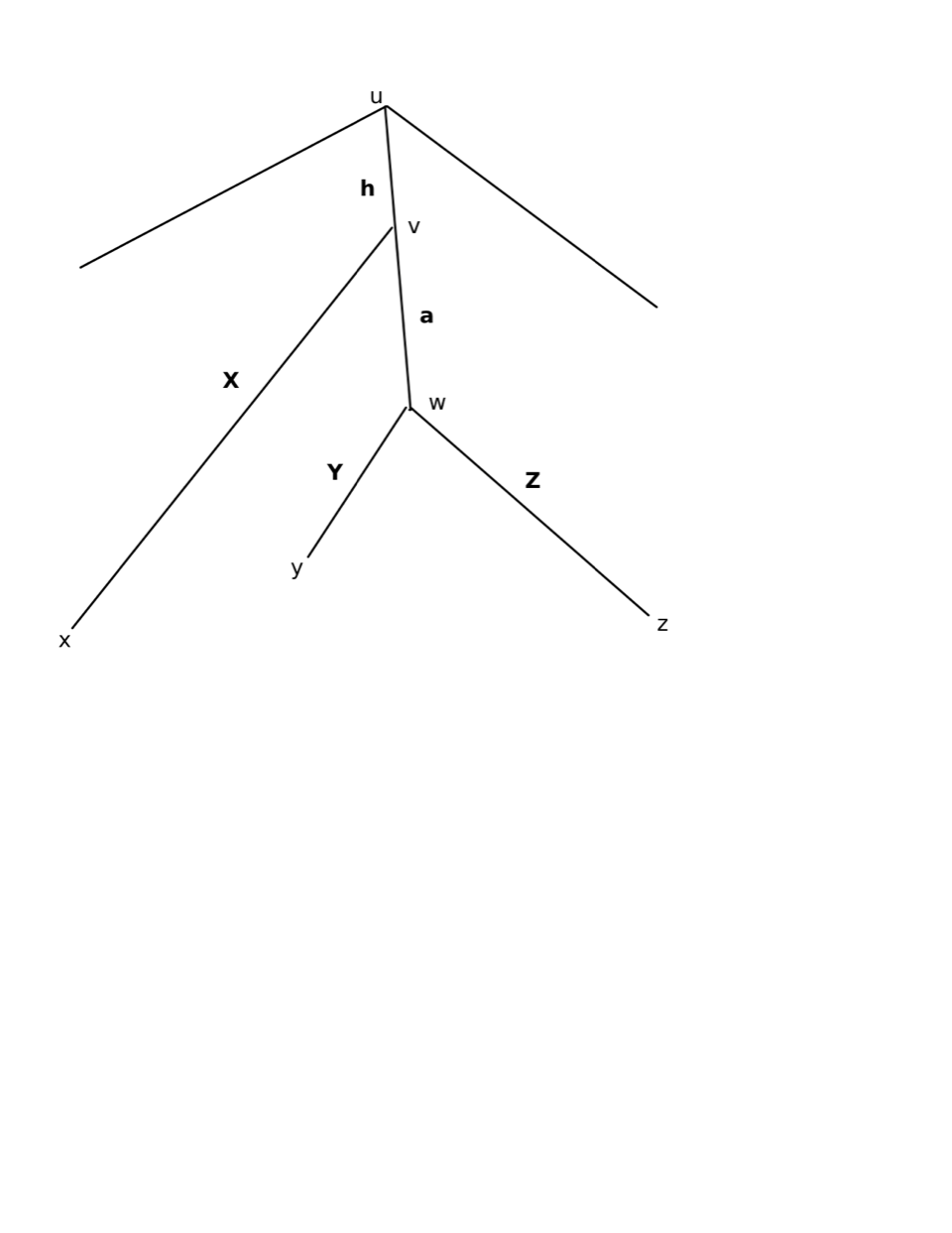}
 \vspace*{-6cm}

 \vspace*{1mm}
\lft Inequality (2): We again proceed as above: we first define the 
following polynomial expressions:

\vspace*{1mm}
\lft $eq3:  (X+Y+2*h+a+2) * (X+Z+2*h+a+2) * (Y+Z+2*h+2*a+2)$;

\lft $eq4:  (h+1) * (Y+Z+2*h+2*a+2) * ( 2*X+Y+Z+4*h+2*a+4)$;

\lft $eq5: (h+a+1) * (X+Y+2*h+a+2) * (X+Z+2*h+a+2)$;

\vspace*{1mm}
\lft If we set \, $eqn: eq3 + 2*eq5 - 2*eq4$, \,  we get 

\vspace*{1mm}
\lft $eqn:  - 2(h+1) * (Z+Y+2h+2a+2) * (Z+Y+2X+4h+2a+4) + \\ \hspace*{6mm}
           (Y+X+2h+a+2) * (Z+X+2h+a+2)(Z+Y+2h+2a+2) + \\ \hspace*{6mm}
             2 (h+a+1) * (Y+X+2h+a+2)* (Z+X+2h+a+2)$
  
 \vspace*{2mm}
 \lft To prove  inequality (2), we need to show that $eqn$ remains non-negative
 for all  non-negative values of $X,Y,Z,h,a$. Expanding $eqn$ (with {\em Maxima}), 
 we  get:

 \vspace*{1mm}
  \lft $eqn$:
  $Y Z^2 + X Z^2 + a Z^2 + Y^2 Z + 2 XYZ + 4 hYZ + 6 aYZ + 4YZ + X^2 Z + 4 hXZ + 6 aXZ + 4 XZ
      + 8 ahZ + 5 a^2Z + 8 aZ + XY^2 + aY^2 + X^2 Y + 4 hXY + 6 aXY + 4XY + 8 ahY + 5 a^2Y 
      + 8 aY+ 4 hX^2 + 4 aX^2 + 4X^2 + 8 h^2X + 16 ahX + 16 hX + 8 a^2X
         + 16 aX + 8 X + 8 ah^2 + 12 a^2h + 16 ah + 4 a^3 + 12 a^2 + 8 a$

   \vspace*{2mm}
  \lft The coefficients are all positive, so we are done.         \hfill $\Box$.

\end{document}